# WHY MACHINE LEARNING INTEGRATED PATIENT FLOW SIMULATION?

*Dr. Tesfamariam M. Abuhay*

University of Gondar
Gondar, Ethiopia
tesfamariam.mabuhay@uog.edu.et

*Dr. Adane L. Mamuye*

University of Gondar
Gondar, Ethiopia
adane.letta@uog.edu.et

*Prof. Stewart L. Robinson*

Loughborough University
Loughborough, UK
S.L.Robinson@lboro.ac.uk

*Dr. Sergey V. Kovalchuk*

ITMO University
Saint Petersburg, Russia
kovalchuk@itmo.ru

**ABSTRACT**

Patient flow analysis can be studied from clinical and/or operational perspective using simulation. Traditional statistical methods such as stochastic distribution methods have been used to construct patient flow simulation sub-models such as patient inflow, Length of Stay (LoS), Cost of Treatment (CoT) and Clinical Pathway (CP) models. However, patient inflow demonstrates seasonality, trend and variation over time.  LoS, CoT and CP are significantly determined by patients' attributes and clinical and laboratory test results. For this reason, patient flow simulation models constructed using traditional statistical methods are criticized for ignoring heterogeneity and their contribution to personalized and value-based healthcare. On the other hand, machine learning methods have proven to be efficient to study and predict admission rate, LoS, CoT, and CP.  This paper, hence, describes why coupling machine learning with patient flow simulation is important and proposes a conceptual architecture that shows how to integrate machine learning with patient flow simulation.

## 1  INTRODUCTION

Patient flow analysis to and in a hospital has been one of the hot research areas in health services research and health economics (Kreindler, 2017; Gualandi et al, 2019). It can be conducted from clinical and/or operational perspective (Côté, 2000) in a single unit/department (e.g., ambulatory care unit (Santibáñez, 2009), intensive care unit (Benjamin and Christensen, 2012), emergency department (Konrad et al, 2013; Cocke et al, 2016; Hurwitz et al, 2014), surgery department (Antonelli et al, 2014; Azari-Rad, 2014)) or in multiple units/departments (e.g., Abuhay et al (2016 and 2020), Kovalchuk et al (2018),  Suhaimi et al (2018)) of a hospital using simulation.

Simulation allows to represent complex systems (Anatoli, 2013) and produces a range of data to support decision making (Monks et al, 2016). Patient flow analysis using simulation can be used to reduce the chances of failure, to meet specifications, to eliminate unforeseen bottlenecks, to prevent under or over-utilization of resources, to reduce crowding, to improve clinical pathways and system performance (Maria and Anu, 1997).  According to Gunal (2012), simulation methods that are employed to study patient flow are generally classified into three categories: Discrete-Event Simulation (DES), Agent-Based Simulation (ABS) and System Dynamics (SD).

To construct components/sub-models (e.g., patient inflow, Length of Stay (LoS), Cost of Treatment (CoT) and clinical pathway (CP) models) of a patient flow simulation, traditional statistical methods such as stochastic distribution (discrete and continues) methods have been used. However, patient inflow or admission rate data demonstrate seasonality and trend and it also varies from hour to hour, day to day, week to week, month to month and year to year (Nas and Koyuncu, 2019). This makes modelling patient admission rate using stochastic (discrete and/or continues) distribution difficult. LoS, CoT and CP are also significantly determined by a patient's attributes such as age, gender, comorbidity, genomic makeup and clinical and laboratory test results (Bramkamp, 2007; Noohi et al, 2020; Siddiqui et al, 2018; Zhang et al, 2010). For this reason, patient flow simulation models are criticized for ignoring heterogeneity (Zaric, 2003) and their contribution to personalized medicine (Schleidgen et al, 2013) and value-based healthcare (Brown, 2005; Traoré, 2019) is now in question.

On top of that, decision makers have a doubt on validity and credibility of patient flow simulation due to significant uncertainty in the patient flow simulation models (Kovalchuk et al, 2018). This, in turn, affects acceptance level and applicability of patient flow simulation models. Patient flow simulation thus needs an accurate estimation model of patient arrival, LoS, CoT and CP (Nas and Koyuncu, 2019).

On the other hand, Electronic Health Record (EHR) (Ambinder, 2005) presents an opportunity by generating big data that can be employed to construct data-driven clinical and/or operational decision support tools that facilitate modelling, analysing, forecasting and managing healthcare.

Machine Learning (ML) (Ngiam and Khor, 2019), using EHR data as an input, has been widely used to study, discover patterns and predict patients' admission rate or demand for healthcare (Asheim, 2019; Luo, 2017; Hong, 2018), LoS (Daghistani, 2019; Taleb, 2017), CoT (Bremer et al, 2018; Jödicke et al, 2019), and CP (Kovalchuk et al 2018; Allen et al, 2019; Prokofyeva and Zaytsev, 2020; Funkner, 2017), to mention a few.

This paper, hence, aims to describe why coupling machine learning with patient flow simulation is important and proposes a conceptual framework that shows how to integrate machine learning with patient flow simulation.

The proposed architecture may improve credibility and acceptance of patient flow simulation as it expands the knowledge stock of general and domain-specific conceptual modelling (Robinson, 2020) of patient flow simulation. It may also foster personalized medicine and value-based healthcare because both concepts promote individual-patient-based healthcare with high-quality, low cost and wide access instead of "one-model-fits-all" approach (Schleidgen et al, 2013; Brown, 2005; Traoré, 2019).

The rest of the paper is organized as follows: Section 2 discusses related works, Section 3 presents why machine learning integrated patient flow simulation is important, Section 4 illustrates conceptual architecture of the proposed model and Section 5 presents conclusion.

## 2     RELATED WORKS

Several studies employed computer simulation methods such as SD, DES, and ABS for modelling patient flow in a single or multiple departments. However, there is still a high variation of uncertainty in patient flow (Kovalchuk et al, 2018) because the performance of patient flow models highly depends on input variables/data.

To minimize the uncertainties and improve accuracy of patient flow simulation, data-driven methods were proposed to supplement the existing simulation modelling techniques. For example, Kovalchuk et al (2018) studied simulation of patient flow in multiple healthcare units using process and data mining techniques for model identification. ML was applied to identify classes of clinical pathways, capturing rare events and variation in patient.

Nas and Koyuncu (2019) proposed an Emergency Department (ED) capacity planning using a Recurrent Neural Network (RNN) and simulation approach. The main objective of this study was to determine patients' arrival times and optimum number of beds in an ED by minimizing the patients' LoS. The outcomes of ML model, hourly patient arrival rates prediction model, was used as input variables.

However, these studies inadequately described why machine learning integrated patient flow simulation was important. Previous studies also did not propose an architecture that shows how to couple machine learning with patient flow simulation.

## 3 WHY MACHINE LEARNING INTEGRATED PATIENT FLOW SIMULATION?

The patient flow simulation model mainly contains two sub-models: patient inflow simulation model and in-hospital patient flow simulation model. The first sub-model simulates patients' arrival to a specific department of a hospital in hourly, daily, weekly or monthly period. Discrete stochastic distribution methods, mainly Poisson distribution (Banks, 2005), were used to develop patient admission rate simulation model. However, patient inflow data exhibits trend, seasonality and/or variation as shown in Figure 1 and Figure 2 (All Figures in this paper were generated using the Acute Coronary Syndrome (ACS) patients' data collected from the Almazov National Medical Research Centre[1], Saint Petersburg, Russia).

Zhang et al (2020) investigated emergency patient flow forecasting in the radiology department and their result implied that ward patient visits had significant nonlinear trend. i.e., the patient arrival problems are generally related hourly, daily, weekly, or monthly. This significantly affect planning and allocation of resources such as bed, health professionals, and diagnosis and treatment equipment.

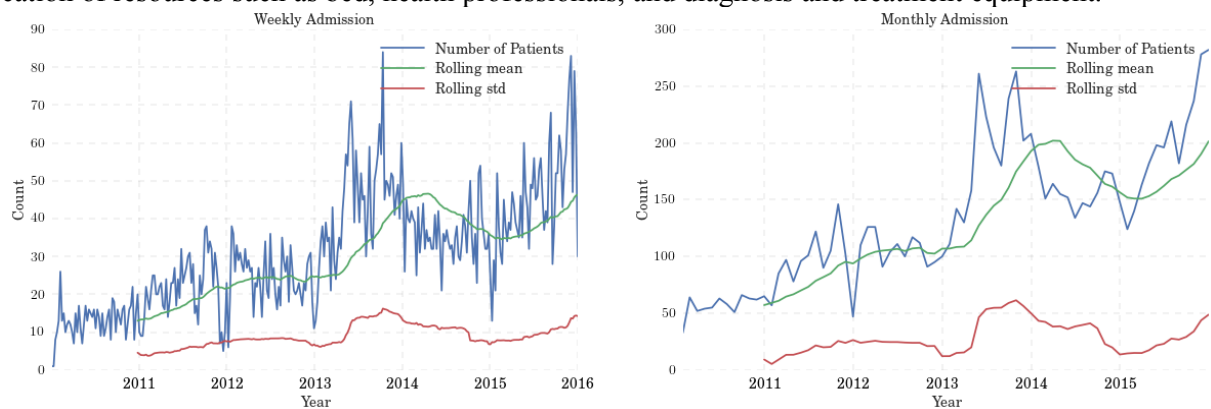

**Figure 1** *Trend and Seasonality of Weekly and Monthly Patient Inflow*

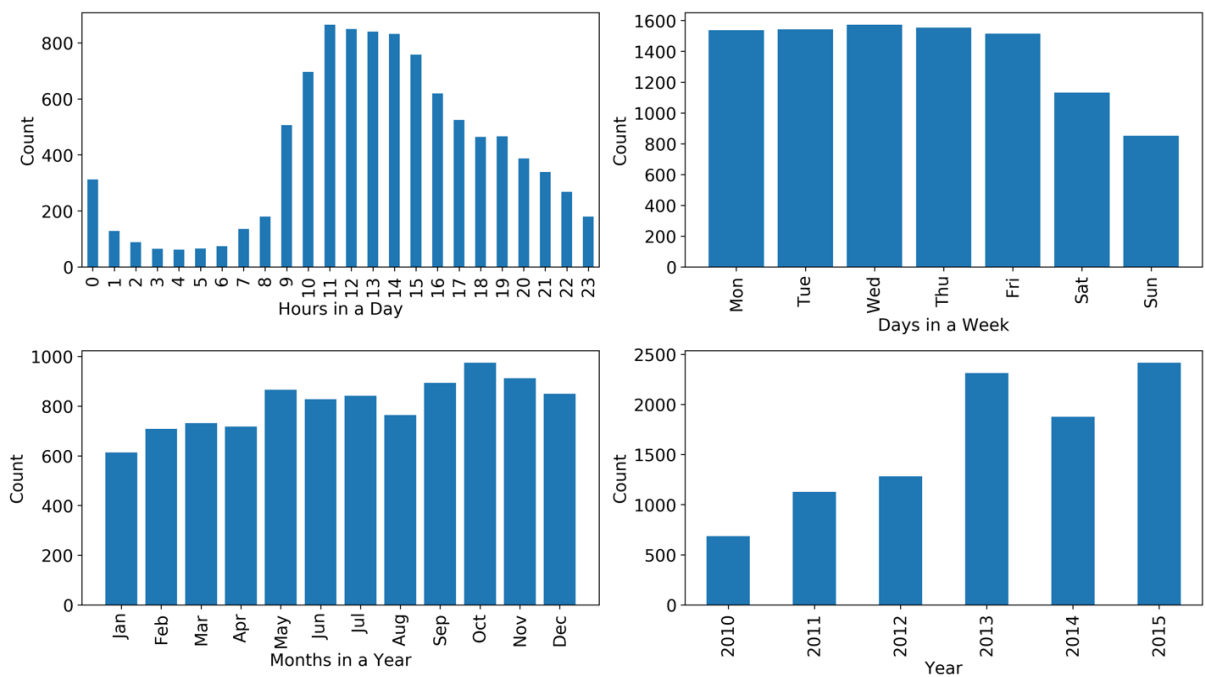

**Figure 2** *Variation of Patient Inflow over Hours, Week Days, Months and Years*

So as to provide timely, personalized and value-based healthcare, understanding the trend of patient flow to a hospital and develop a prediction model that can identify and understand trend, seasonality and variation of patient flow is vital. Hence, instead of modelling patient inflow with stochastic distribution (counting functions) methods, ML algorithms for time series analysis and prediction (Brockwell, 2010) are ideal because they are able to pinpoint seasonality and trend and predict the future patient inflow, in different time period, based on historical data. Woo et al (2018) also showed that comprehensive elements of patient history (e.g., previous healthcare usage statistics, past medical history, historical labs and vitals, prior imaging counts, and outpatient medications, and demographic details such as insurance and employment status) improved patient inflow prediction performance significantly.

---

[1] http://www.almazovcentre.ru/?lang=en

Machine learning, therefore, would have twofold objectives: simulating patient inflow and/or predicting future patient inflow. This allow effective and efficient planning and resources allocation, while reducing over- and/or under-utilization.

The second sub-model, which is in-hospital patient flow simulation model, mimics movement of patients through multiple clinical and/or operational processes in a single or multiple departments. In the case of multiple departments, the in-hospital model simulates movement of patients from one department to another based on transaction matrix. This model may have sub-models such as LoS estimation model, CoT estimation model and CP estimation model.

The LoS is a significant indicator of the effectiveness and efficiency of a hospital management. LoS has been used as a surrogate to evaluate the utilization of resources, quality and efficiency of care, costs of treatment, patient experience, and planning capacity in a hospital (Papi, 2016; Verburg et al, 2014). Both LoS and CoT sub-models are usually developed using univariate density estimation methods such as Lognormal, Weibull, and Gamma. However, (Ickowicz et al, 2016; Lee et al, 2011; Houthooft et al, 2015) mentioned that none of them seemed to fit satisfactorily in a wide variety of samples.

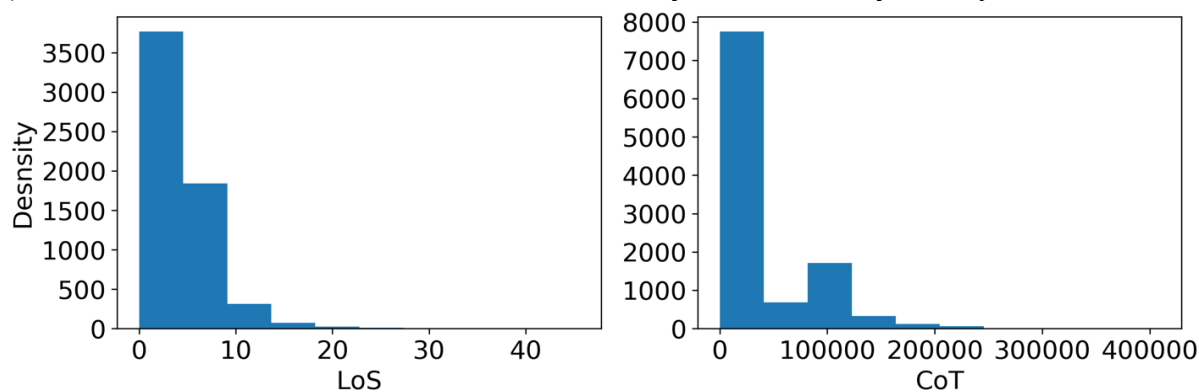

**Figure 3** *Distribution of Length of Stay (LoS) in Days and Cost of Treatment (CoT) in RUB*[2]

Instead, the assumption of heterogeneous sub-populations would be more appropriate because the probability distribution of both LoS and CoT is positively skewed, multi-modal (See Figure 3) and significantly vary between diagnosis-related groups (DRGs) and correlated with characteristics of patients such as age, gender, number of comorbidity (see Figure 4) and the like (Daghistani et al, 2019; Bramkamp et al, 2007; Noohi et al, 2020; Siddiqui et al, 2018; Ickowicz et al, 2016). This limits the use of inference techniques based on normality assumptions (Ickowicz et al, 2016).

According to Ngiam and Khor (2019), ML provides flexibility and scalability compared with traditional statistical methods because it allows to analyse diverse data types and incorporate them into predictions for disease risk, diagnosis, prognosis, appropriate treatments, LoS and CoT. Thus, modelling LoS and CoT using ML while constructing patient in-hospital flow simulation allow considering heterogeneous sub-populations based on their characteristics. This may improve accuracy and credibility of patient flow simulation model.

---

[2] Russian Federation Currency

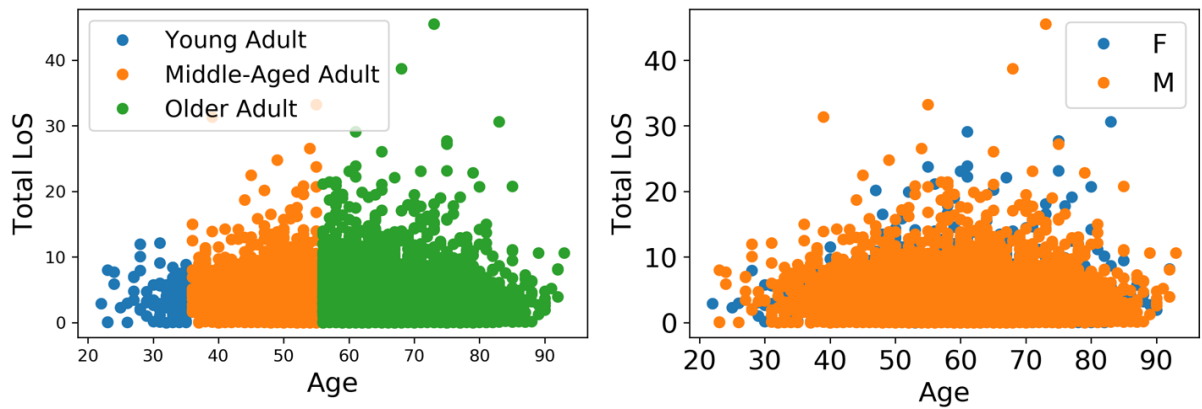

**Figure 4** *Distribution of LoS based on Age Category and Gender*

Another sub-model of patient flow simulation is CP prediction model. CP specify the categories of care, activities, and procedures that need to be conducted for a group of patients until they are discharged from the hospital (Aspland et al, 2019). Modelling the processes in a healthcare system plays a large role in understanding its activities and serves as the basis for increasing the efficiency of medical institutions (Prokofyeva and Zaytsev, 2020).

Clinical pathways have been generally modelled based on probability law using transaction matrix. However, patient in-hospital flow is very complex process due to dissimilar and multiphase pathways and the innate uncertainty and variability of care processes due to patients' attributes and their previous history. On top of this, CP identification and prediction involves analysis of comprehensive patient information (Kovalchuk, 2018; Aspland et al, 2019) that cannot be achieved or modelled with transaction matrix. Hence, pathway analysis and prediction model using ML while constructing in-hospital patient flow simulation may foster personalized medicine and improve efficiency of healthcare provision.

## 4 HOW TO INTEGRATE MACHINE LEARNING WITH PATIENT FLOW SIMULATION?

Figure 5 shows how machine learning can be coupled with patient flow simulation. Electronic medical record (EHR), a digital version of a patient's paper chart, generates huge amount of data that can be used to construct data-driven decision support tools that facilitate modeling, analyzing, forecasting and managing operational and/or clinical processes of healthcare.

Different kind of data about admission rate, LoS, CoT, characteristics of patients (e.g., age, gender, genetic makeup and etc.), clinical pathways in the form of event log, clinical and laboratory test results can be extracted from the EHR. This would allow to develop data-driven patient flow simulation (Ambinder, 2005).

Instead of modelling patient inflow using stochastic distribution, it can be formulated as time series problem and modelled using ML algorithms for times series problem. So that factors affecting patient flow to a hospital can be considered and seasonality and trending nature of patient inflow can be captured.

Patient inflow can be modelled as a univariate or multivariate data. In the case of modelling patient inflow as univariate data, hourly, daily, weekly, monthly or yearly number of patients only is extracted and used as an input. Whereas, patient inter-arrival rate (hourly, daily, weekly, monthly or yearly) can be integrated with third-party data such as weather, demographic structure of a population, pandemic and natural and/or human made disasters and used as an input to model patient inflow as multivariate data. Data integration and preprocessing tasks can be applied so as to handle missing values, normalize the data and make the data suitable for ML algorithms.

The patient inflow prediction model can be developed for hourly, daily, weekly, monthly or yearly period. Zhang et al (2020) predicted emergency patient flow in the radiology department by constructing six linear (autoregressive integrated moving average and least absolute shrinkage and selection operator) and nonlinear models (linear-and-radial support vector regression models, random forests and adaptive boosting) and considering the lag effects and corresponding time factors. The data was collected from the radiology department and the performance of the models was measured using mean absolute percentage error.

Khaldi et al (2019) attempted to forecast weekly patient visits to ED by combining Artificial Neural Networks (ANNs) with a signal decomposition technique named Ensemble Empirical Mode decomposition (EEMD). Seven years of univariate time series data of weekly demand was collected from ED and the proposed models were evaluated using root mean square error (RMSE), mean absolute error (MAE) and correlation coefficient (R).

The ML-based patient inflow prediction model is the starting point of the patient flow simulation process. It can be also attached to each department so that demand and supply can be analyzed, forecasted and managed at department or operational level, if the simulation model encompasses multiple departments/units.

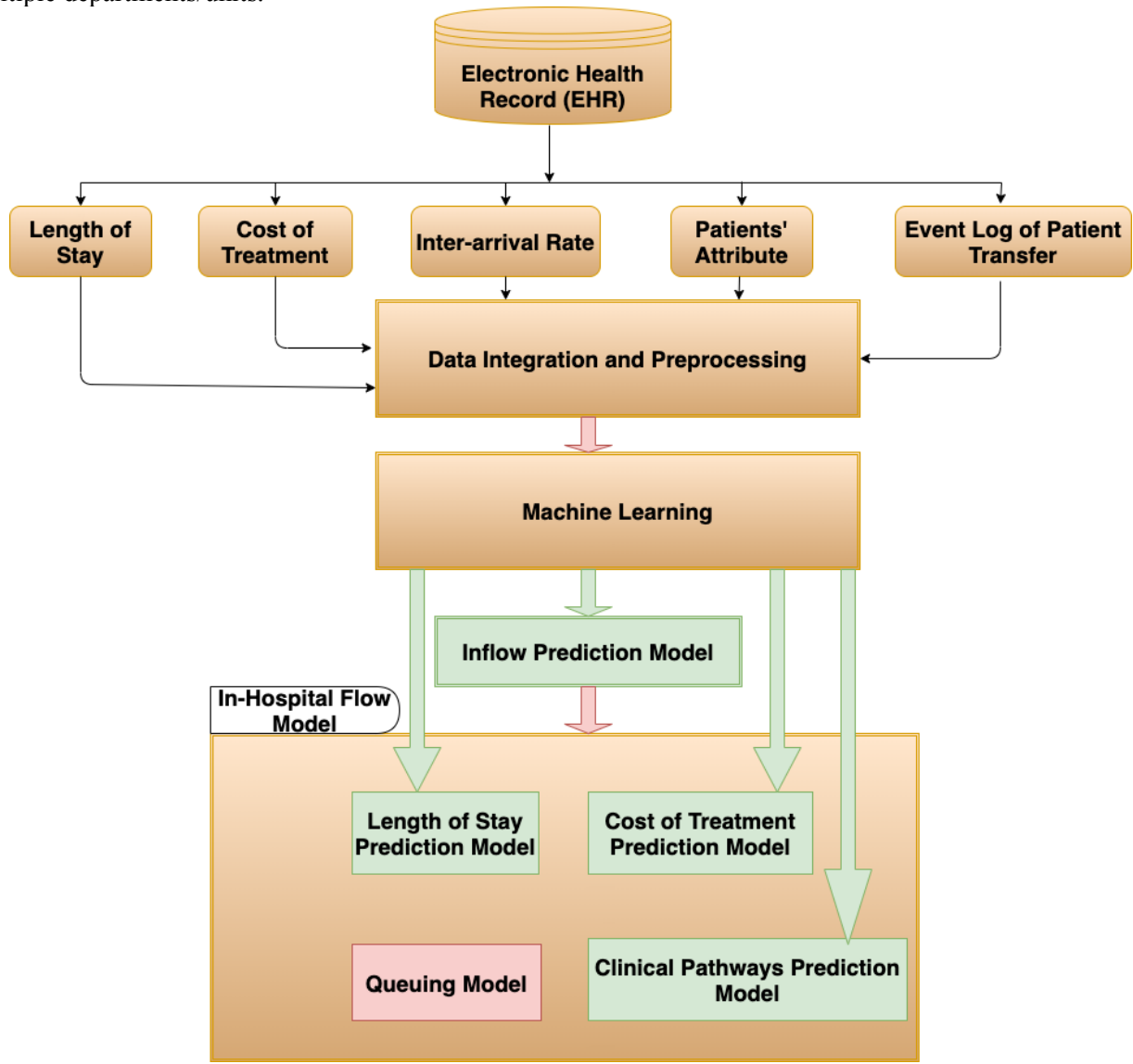

**Figure 5** *Conceptual architecture of the proposed model*

Different factors, including characteristics of patients, affect LoS, CoT and CP. Hence, formulating LoS and CoT as regression problem allow consideration of different determinate factors such as characteristics of patients, clinical and laboratory tests results and co-morbidities. So that sub-populations of patients could be managed accordingly and unnecessary LoS can be reduced, which results in decreased risk of healthcare acquired infection, improvement of quality of treatment, reduction of CoT, and increased availability of free beds for needy patients (Baek et al, 2018).

Linear and nonlinear models can be used to model and predict LoS. Baek et al (2018), for example, analyzed LoS using electronic health records and machine learning techniques of multiple regression analysis. Muhlestein et al (2019) trained 29 ML algorithms, including tree-based models, linear classifiers, support vector machines, neural networks, and naïve Bayes classifiers, on 26 preoperative variables, collected from publicly available NIS database, to predict LoS following craniotomy for brain tumor. The root mean square logarithmic error (RMSLE) was applied to evaluate the performance of the ML and top performing algorithms were combined to form an ensemble. Mekhaldi et al (2020) compared two ML methods, the Random Forest (RF) and the Gradient Boosting model (GB), to predict the LoS based on an open source dataset and the Mean Square Error (MAE), the R-squared ($R2$) and the Adjusted R-squared (Adjusted $R2$ ) metrics were used to evaluate the performance of the models. The same fashion can be also applied to analyze and model CoT as both LoS and CoT demonstrate similar behavior.

The ML-based LoS and CoT prediction models will be attached to each department or healthcare service in a department so that demand and supply can be analyzed, forecasted and managed at operational level. In order to attach LoS and CoT prediction model to each department/healthcare services in the in-hospital patient flow simulation, generator of characteristics of patients should be constructed using continues and/or discrete probability functions according to the data type of the patients' characteristics.

Clinical pathways or patient trajectories analysis and prediction can be formulated as a clustering problem using clustering methods. Event log data about movement of patients, in combination with patient characteristics, previous clinical history, current laboratory and clinical test results, can be used to model and predict clinical pathways in a single department or in the entire hospital. Prokofyeva and Zaytsev (2020) analyzed clinical pathways in medical institutions using hard and fuzzy clustering methods and public data. Allen et al (2019) studied significance of machine learning to analyze clinical pathway and enhance clinical audits. Secondary data collected during routine care were used, following a comparison of methods, a random forests method was chosen and the model was validated using stratified tenfold validation. Allen et al (2019) proposed data-driven modeling of clinical pathways using EHR to cluster patients into groups based on their movements/clinical pathways during their stay in hospital. Kovalchuk et al (2018) studied simulation of patient flow in multiple healthcare units using process and data mining techniques for model identification. ML was applied to identify classes of clinical pathways, capturing rare events and variation in patient using clustering.

## 5   CONCLUSION

The aim of this paper is to describe why machine learning integrated patient flow simulation is important and to propose a conceptual architecture that shows how to integrate machine learning with patient flow simulation.

Traditional statistical methods such as stochastic distribution (discrete and continues) methods have been used to construct sub-models (e.g., patient inflow, Length of Stay (LoS), Cost of Treatment (CoT) and clinical pathways models) of patient flow simulation model. However, patients' admission data demonstrate seasonality, trend and variation over time. LoS, CoT and clinical pathways are also significantly determined by a patient's attributes such as age, gender, comorbidity, genomic makeup and clinical and laboratory test results. For this reason, patient flow simulation models were criticized for ignoring heterogeneity and their contribution to personalized medicine and value-based healthcare is now in question.

On the other hand, machine learning methods have proven to be efficient to study and predict patients' admission rate, bed capacity, LoS, CoT, and clinical pathways. This paper, hence, describes why coupling machine learning with patient flow simulation is important and proposes a conceptual architecture that shows how to integrate machine learning with patient flow simulation models.

**ACKNOWLEDGMENTS**

The authors thank Almazov National Medical Research Centre (Saint Petersburg, Russia) for providing anonymized data for this study.

## AUTHOR BIOGRAPHIES

**Tesfamariam Abuhay** received a BSc in Management Information Systems (MIS) from Haramaya University of Ethiopia in 2010. He completed his MSc in IT at the University of Gondar, Ethiopia in 2015 and his PhD in Computer Science at ITMO University, Russia in 2019. He is currently an Assistant Professor and the Vice Dean of College of Informatics at the University of Gondar, Ethiopia. Dr Abuhay has an extensive background in data-driven patient flow analysis and simulation for the past 5 years. http://www.uog.edu.et/foi/personnal/dr-tesfamariam-mulugeta/

**Adane Mamuye** is an Assistant Professor at the College of Informatics at the University of Gondar, Ethiopia. He is and project coordinator of Capacity Building Mentorship Program (CBMP) and he is also Co-Director of eHealth Lab Ethiopia. He received his Ph.D from University of Camerino, Italy, in 2017. https://ehealthlab.org/our-team/adane-l-mamuye/

**Stewart Robinson** is a Professor of Management Science and Dean of School of Business and Economics at Loughborough University, UK. He completed his BSc and PhD in Management Science (Operational Research) from Lancaster University and was Fellow of the Operational Research Society. http://www.stewartrobinson.co.uk/

**Sergey Kovalchuk** is an Associate Professor and a senior researcher at the eScience Research Institute and a lecturer at the High-Performance Computing Department at ITMO University. His education includes a degree in software engineering from Orenburg State University (Orenburg, Russia) in 2006 and a PhD degree (Candidate of Technical Sciences) in the field of mathematical modelling, numerical calculations, software systems from ITMO University (Saint Petersburg, Russia) in 2008 (thesis subject "High-Performance Software System for Metocean Extreme Events Simulation"). https://en.itmo.ru/en/viewperson/1248/Sergey_Kovalchuk.htm